%% file: NAACL2019Discourse.tex
\documentclass[11pt,a4paper]{article}
\usepackage[hyperref]{naaclhlt2019}
\usepackage{booktabs} 
\usepackage{graphicx}  
\usepackage{booktabs} 
\usepackage{amsmath}
\usepackage{amsfonts}
\usepackage{comment}
\usepackage{footmisc}
\usepackage{mathrsfs}
\usepackage{balance}
\usepackage{amssymb}
\usepackage{multirow}
\usepackage{multicol}  
\usepackage{tikz}
\usepackage{enumitem}
\usepackage{array}
\usepackage{float}
\usepackage{caption}
\usepackage{cleveref}
\usepackage{hyperref}
\usepackage{caption}
\usepackage{subcaption}
\usepackage{xcolor}
\usepackage{perpage}
\MakePerPage{footnote}
\usepackage[ruled,linesnumbered]{algorithm2e}
\usepackage{thmtools}
\usepackage{algpseudocode}

\SetCommentSty{mycommfont}
\declaretheorem[name=Property]{Property}

\aclfinalcopy 

\title{Learning Hierarchical Discourse-level Structure for Fake News Detection}
\author{Hamid Karimi \\
  Computer Science and Engineering\\
  Michigan State University \\
  {\tt karimiha@msu.edu} \\\And
  Jiliang Tang  \\
   Computer Science and Engineering\\
  Michigan State University \\
  {\tt tangjili@msu.edu}
}
\date{}

\begin{document}
\maketitle

\begin{abstract}

On the one hand, nowadays, fake news articles are easily propagated through various online media platforms and have become a grand threat to the trustworthiness of information. On the other hand, our understanding of the language of fake news is still minimal. Incorporating hierarchical discourse-level structure of fake and real news articles is one crucial step toward a better understanding of how these articles are structured. Nevertheless, this has rarely been investigated in the fake news detection domain and faces tremendous challenges. First, existing methods for capturing discourse-level structure rely on annotated corpora which are not available for fake news datasets. Second, how to extract out useful information from such discovered structures is another challenge. To address these challenges, we propose  \textbf{H}ierarchical \textbf{D}iscourse-level \textbf{S}tructure for \textbf{F}ake news detection. HDSF learns and constructs a discourse-level structure for fake/real news articles in an automated and data-driven manner.
Moreover, we identify insightful structure-related properties, which can explain the discovered structures and boost our understating of fake news. Conducted experiments show the effectiveness of the proposed approach. Further structural analysis suggests that real and fake news present substantial differences in the hierarchical discourse-level structures.

\end{abstract}

\input{introduction.tex}
\input{problem.tex}
\input{model.tex}

\input{structure-property.tex}

\input{experiments.tex}
\input{comparison.tex}
\input{structure-exp.tex}
\input{related.tex}

\section{Conclusion and Future Work}
\label{sec:conclusion}

In this work, we looked into fake news detection from a new perspective.
We hypothesized that hierarchical discourse-level structure of news documents offers a discriminatory power for fake news detection. To investigate this hypothesis, we proposed a new framework HDSF, which can automatically extract discourse-level structures of real/fake news documents represented by dependency trees while does not rely on an annotated corpus. Moreover, we defined a set of insightful properties describing tree structures. Conducted experiments confirmed the power of our approach where it outperformed representative baselines. More importantly, we highlighted noticeable differences between structures of fake and real news documents. These differences also indicated less coherency in the fake news documents.

The new perspective pursued in this paper can be continued in several directions. First, we intend to define more advanced properties from the discourse dependency trees.  Second, investigating the hierarchical structure at the word-level will be an exciting research inquiry. Finally, unsupervised discourse-level structure extraction of fake/real news documents is a worthwhile research topic. 

\section*{Acknowledgment}

This work is supported by the National Science Foundation (NSF) under grant numbers IIS-1714741, IIS-1715940, IIS-1845081, CNS-1815636, and a grant from Criteo Faculty Research Award. We would like to thank Mr. Tyler Derr and other members of the Data Science and Engineering (DSE) lab at Michigan State University\footnote{\url{http://dse.cse.msu.edu/}} for their constructive comments.   

\balance
\bibliography{Tree-Reference}
\bibliographystyle{acl_natbib}

\end{document}

%% file: introduction.tex
\section{Introduction}
\label{sec:introduction}
In this work, we focus on detecting fake news articles (hereafter referred to as \emph{documents}) based on their contents. Many existing linguistic approaches for fake news detection~\cite{Feng-etal2012, Pennebaker-etal2015, Ott-etal2011} overlook a crucial linguistic aspect of fake/real news documents i.e., the hierarchical discourse-level structure. Usually, in a document, discourse units (e.g., sentences) are organized in a hierarchical structure e.g., a tree. The importance of considering the hierarchical discourse-level structure for fake news detection is three-fold. First, previous studies~\cite{Bachenko-etal2008,Rubin-Lukoianova2015} explored discourse-level structure in fake news detection and discovered that the way two discourse units of a document are connected could be quite revealing and insightful about its truthfulness. For instance, \cite{Rubin-Lukoianova2015} applied Rhetorical Structure Theory (RST)~\cite{Mann-Thompson1988} and noted that fake stories lack ``evidence" as a defined inter-discourse relation. Second, fake news is typically produced by connecting disjoint pieces of news and unlike well-established journalism (e.g., New York Times) fake news production lacks a meticulous editorial board. Therefore, by incorporating the hierarchical discourse-level structure, we can investigate the coherence of fake/real news documents (we will show this later). Third, a substantial number of studies have shown that using hierarchical structures yields a better document representation in various downstream tasks whose predictions depend on the entire text~\cite{bhatia2015better,Mathieu-etal2018,Li-etal2014b}. Since typically fake news detection is considered as a classification problem based on the entire text, applying discourse analysis has the potential to advance fake news detection (this will be verified later). 

On the other hand, incorporating the hierarchical structure at the discourse level for fake news detection faces tremendous challenges. First,  many existing methods incorporating structural discourse~\cite{Li-etal2014,bhatia2015better} (not for fake news detection though) rely on annotated corpora such as Penn Discourse Treebank~\cite{Prasad-etal2007}. Constructing and annotating such corpora is an arduous and costly process. Incorporating hierarchical structure is even more difficult for fake news detection as there exists virtually no available annotated discourse corpus in this domain. Therefore, we need to learn the discourse-level structure in a data-driven and automated manner. 
Second, how to use the hierarchical discourse-level structure and extract insightful information that can boost our understanding of fake news is another challenge.

\begin{figure}
\centering
\includegraphics[width=\columnwidth]{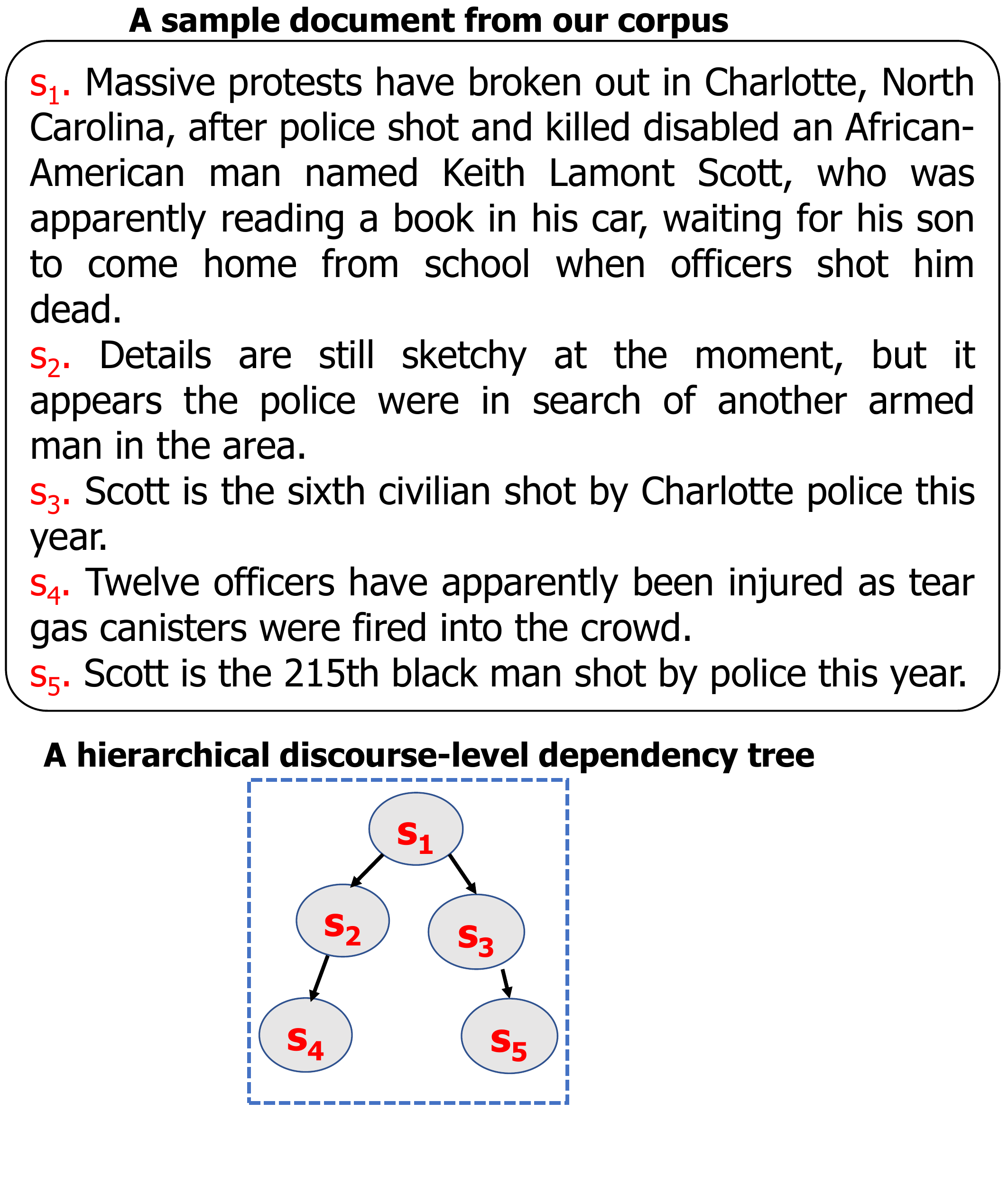}
\caption{An illustration of the hierarchical discourse-level structure of a document using a dependency tree}
\label{fig:dep-tree-example}
\end{figure}

In this study, we embrace the opportunities and challenges and propose \textbf{H}ierarchical \textbf{D}iscourse-level  \textbf{S}tructure for \textbf{F}ake news detection (HDSF) framework. HDSF in an automated manner learns a hierarchical structure for a given document through an approach based on the dependency parsing at the sentence level (i.e., sentences are discourse units). As an example, Figure~\ref{fig:dep-tree-example} illustrates a hierarchical structure of a document i.e., a dependency tree of sentences. Sentences represent discourse units and two sentences are connected by a directed link where a sentence at the \emph{head} of the link  \textit{semantically depends} on a sentence at the \emph{tail} of the link e.g., in Figure~\ref{fig:dep-tree-example}, the sentence $s_4$ depends on the sentence $s_2$.    

Our key contributions are summarized as follow.
\begin{itemize}
\item To the best of our knowledge, we are the first to study automatic document structure learning for fake news detection. 
\item We propose the framework HDSF that automatically and in an end-to-end manner learns structurally rich representations for fake and real news documents.
\item We identify a set of structure-related properties delineating meaningful structural differences between fake and real news documents.
\end{itemize}

The rest of the paper is organized as follows. In Section~\ref{sec:problem}, we formally define the problem. Section~\ref{sec:model} describes the proposed framework. In Section~\ref{sec:properties}, we introduce several structure-related properties to analyze the hierarchical structures. Section~\ref{sec:experiments} presents the experiments and discussions followed by an overview of the related work in Section~\ref{sec:related}. We conclude the paper in Section~\ref{sec:conclusion} and shed light on a few future directions. 

%% file: problem.tex
\section{Notations and Problem Statement}
\label{sec:problem}
 Following the previous work~\cite{Allcott-Gentzkow2017,Shu-etal2017b}, we define the fake news as follows.

\textbf{Definition}. We define a news document \textit{fake} if its content is verified to be false and \textit{real} otherwise. 

Let's briefly introduce some notations. Suppose we have a corpus $\mathcal{D}$ of fake and real news documents. Let a document $d \in \mathcal{D}$ contain $k$ sentences $s_1, s_2, \cdots s_k$. Suppose a sentence $s_j \in d$ ($1\leq j \leq k$) includes words $W_j =\{w_{1},w_{2},\cdots,w_{T_{j}}\}$ where $T_{j}$ denotes the number of words in sentence $s_j$. Additionally,  binary labels ${Y}$ (i.e., \emph{fake} or \emph{real} labels)  hold ground-truth labels associated with documents in $\mathcal{D}$.

\textit{Given the corpus of fake/real news documents (i.e., $\mathcal{D}$), we aim to learn model  $\mathcal{M}$ that can automatically learn hierarchical and structurally rich representations for documents in $\mathcal{D}$. Meanwhile, given binary labels ${Y}$, the model $\mathcal{M}$  uses the hierarchical representations to automatically predict the labels of unseen news documents.}

%% file: model.tex
\section{The Proposed Framework}
\label{sec:model}

\begin{figure*}
\centering
\includegraphics[scale=0.29]{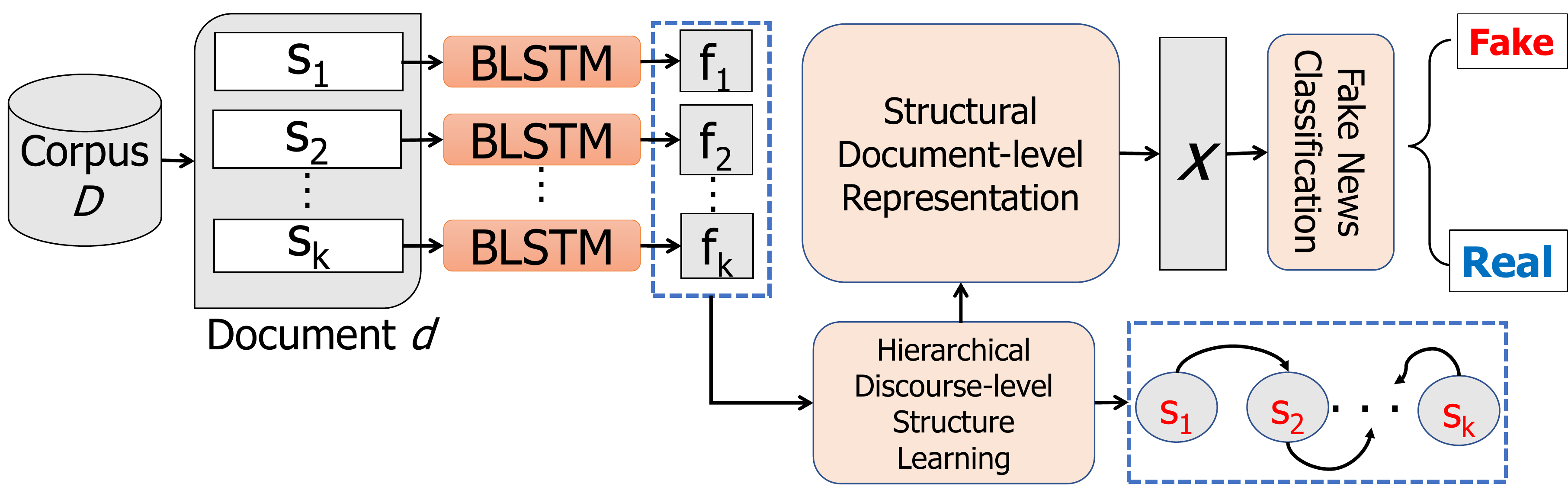}
\caption{The proposed framework Hierarchical Discourse-level  Structure for Fake news detection (HDSF)}
\label{fig:model}
\end{figure*}

To incorporate the discourse-level structure for fake news detection, we propose the framework HDSF illustrated in Figure~\ref{fig:model}. It provides three components: the \emph{Hierarchical Discourse-level Structure Learning}  automatically learns a proper structure for a given document, the \emph{Structural Document-level Representation} yields a numerical and unified representation for the entire document, which is used by the \emph{Fake News Classification} component to predict the label of a document. 

\subsection{Hierarchical Discourse-level Structure Learning}
\label{sec:model-struct-extract}
In this component, we aim to construct a hierarchical structure between discourse units (i.e., sentences) without relying on an annotated corpus. 
To achieve this, we use a dependency parsing approach~\cite{Yang-Lapata2017,Li-etal2014,Kim-etal2017} and represent the hierarchical structure of a document as a dependency tree (see Figure~\ref{fig:dep-tree-example} for an example of a dependency tree). In dependency parsing of discourse units, we mainly need to identify if a discourse unit semantically depends on another one. If so, a \emph{parent-child} (or \emph{tail-head}) link in the dependency tree can be established. Therefore, as long as the semantic dependencies between discourse units are identified, we can construct a discourse-level dependency tree without an annotated corpus. In Section~\ref{sec:interdiscourse}, we describe a method to discover the semantic dependencies between discourse units in an automated manner. Afterward, in Section~\ref{sec:tree-const}, we utilize these dependencies to construct the dependency tree of a document.

\subsubsection{Learning Inter-Discourse Dependencies}
\label{sec:interdiscourse}
Since discourse units are defined as sentences, we first need to get a fixed representation for each sentence. To this end, we utilize Bi-directional Long Short-Term Memory (BLSTM) network~\cite{Schuster-Paliwal1997}. We represent each word in a sentence $s_j$  by a fixed-size word embedding, and further the BLSTM network at each time step $t\in[1,T_j]$ executes the following functions\footnote{Note that the backward LSTM (i.e., the lower part of Eq.~\ref{eq:blstm}) takes a sequence of words in a reverse order.}:

\begin{equation}
\begin{aligned}
\overrightarrow{h}_{t} &= \mathcal{F}(\overrightarrow{h}_{t-1};w_{t-1}) \\ 
\overleftarrow{h}_{t} &= \mathcal{F}(\overleftarrow{h}_{t-1};w_{T_{j}-t+1})
\end{aligned}
\label{eq:blstm}
\end{equation}

\noindent where $\mathcal{F}$ is the LSTM function~\cite{Hochreiter-Schmidhuberl1997}, and $\overrightarrow{h}_t$ and $\overleftarrow{h}_{t} $ are outputs of the forward and backward LSTM networks at time step $t$, respectively.  Then, a fixed representation for a sentence $s_j$, denoted as $f_j$, is defined as the average of the last output of forward and backward LSTM networks:
\begin{equation}
\begin{aligned}
f_j =  \frac{[\overrightarrow{h}_{T_{j}}+\overleftarrow{h}_{T_{j}}]}{2} 
\end{aligned}
\end{equation}

Similarly, we apply the BLSTM network to all sentences of a document  and obtain a sequence of sentential representations i.e., $f_1,f_2,\cdots,f_{k}$ (see Figure~\ref{fig:model}). 

As mentioned before, in dependency parsing, we need to identify the dependency between two discourse units in an automated manner. To do this, the HDSF framework learns and optimizes an inter-sentential attention matrix $\mathbf{A} \in \mathbb{R}^{k\times k}$. The entry $(m,n)$ of $\mathbf{A}$ holds the probability of the sentence $s_m$ being the parent of the sentence $s_n$ where $1 \leq m,n \leq k$ and $m \neq n$. In other words, $\mathbf{A}$ contains parent-child probabilities and is computed as follows.
\begin{equation}
\begin{aligned}
u_m &= \mathcal{G}(\mathbf{W} \times f_m+\mathbf{b})   \\
u_n &= \mathcal{G}(\mathbf{W} \times f_n+\mathbf{b}) \\
\mathbf{A}[m,n] &= \frac{e^{( u_m \odot u_n)}}{\sum_{i=1}^{k} e^{\sum(u_i \odot u_n)} } 
\end{aligned}
\label{eq:sent-attention}
\end{equation}

\noindent  where $\mathcal{G}$ is a non-linear activation function, $\mathbf{W}$ is some weight matrix, $\mathbf{b}$ is a bias vector, and $\odot$ denotes the dot product operator. Further, since we need a root node in a dependency tree,  we compute the probability of a sentence $s_j$ being the root node, denoted as $r_j$, as follows.
\begin{equation}
\begin{aligned}
u_j &= \mathcal{G}(\mathbf{W} \times f_j+\mathbf{b})  \\
r_j &= \frac{e^{\sum_{\forall y} u_j[y]}}{\sum_{i=1}^{k} e^{\sum_{\forall y} u_i[y]}} 
\end{aligned}
\label{eq:atten-root}
\end{equation}
\noindent where $u_j[y]$ is the $y$-th element of vector $u_j$. Similarly, we calculate the root probabilities for all sentences and obtain the array of root probabilities  denoted as $\mathbf{r}= \{r_1, r_2, \cdots, r_k\}$ where $ 0 \leq r_j \in \mathbf{r} \leq 1$.

\subsubsection{Discourse Dependency Tree Construction} 
\label{sec:tree-const}

We use the learned matrix of inter-sentential parent-child probabilities i.e., $\mathbf{A}$ (Eq.~\ref{eq:sent-attention}) as well as the array of root probabilities i.e., $\mathbf{r}$ (Eq.~\ref{eq:atten-root}) and propose a greedy algorithm, illustrated in Algorithm~\ref{alg:tree-const}, to construct the discourse dependency tree of a document. A sentence with the maximum value in $\mathbf{r}$ is considered the root node and is inserted into the tree (line~\ref{alg:root}). Then, at each iteration, the algorithm finds the maximum entry in a block of the matrix $\mathbf{A}$ whose rows correspond to the rows of current nodes added to the tree (i.e., nodes $V$ in line~\ref{alg:pc}) and its columns correspond to columns of the rest of nodes (i.e., nodes $N \backslash V$ in line~\ref{alg:pc}). Note that the columns of the current nodes are excluded because their parents have already been identified and also each node should have exactly one parent (except the root which has no parent). Assume the search in line~\ref{alg:pc} results in the entry $(p,c)$ of $\mathbf{A}$ where $ 1\leq p,c\leq k$ and $p \neq c$. Then, the sentence $s_c$ is added as the child node of the sentence $s_p$ (line~\ref{alg:add2}). Algorithm~\ref{alg:tree-const} continues until all sentences of a document are added to the tree $\mathcal{T}$.

\begin{figure}
\centering
\includegraphics[width=\columnwidth]{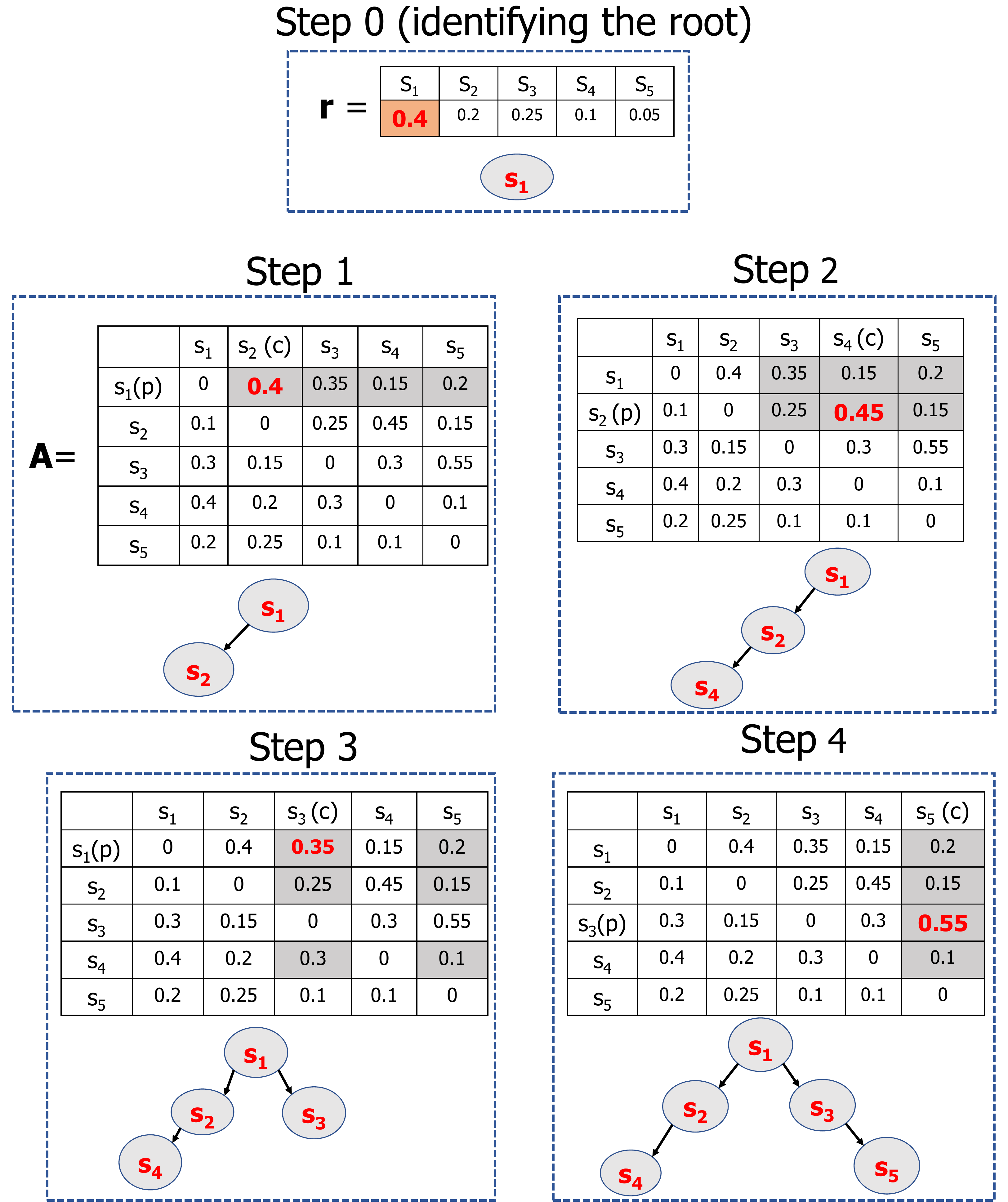}
\caption{An illustration of a step-by-step execution of Algorithm~\ref{alg:tree-const} for the sample document presented in Figure~\ref{fig:dep-tree-example}. The search block at each iteration has been highlighted. Note the letter `p' for the \emph{parent node} and `c' for the \emph{child node}.}
\label{fig:dep-tree-const}
\end{figure}

\begin{algorithm}\small
\caption{The proposed algorithm for discourse dependency tree construction }
\label{alg:tree-const}
    \KwIn
    {
        $\mathbf{A}$; $\mathbf{r}$
        
    }
    \KwOut
    {
        Discourse dependency tree $\mathcal{T}$
    }
    {
        $\mathcal{T}=empty$ \label{alg:T} \\
        $N=\{s_1,s_2,\cdots,s_{k}\}$ \tcp{All nodes}    \label{alg:V} 
        $V=\{\}$ \tcp{Set of current nodes} 
        Add $N[argmax(\mathbf{r})]$ to $V$ \label{alg:add1}  \\
        $\mathcal{T}.root=N[argmax(\mathbf{r})]$ \tcp{Adding the root } \label{alg:root}
    }
    \While{$ |V| \neq k  $\label{alg:loop}} 
    {
        $p,c=argmax(\mathbf{A}[V,N \backslash V])$ \tcp{Search block} \label{alg:pc}
          $\mathcal{T}.link(N[c],N[p])$ \tcp{Child-parent link} \label{alg:add2}
         $V.add(N[c])$ \tcp{Adding the child node}
           
    }
\Return $\mathcal{T}$
\end{algorithm}
To fix the idea of discourse dependency tree construction algorithm, we present a step-by-step execution of this algorithm demonstrated in Figure~\ref{fig:dep-tree-const}. In Step 0, the sentence $s_1$ is added as the root of the tree $\mathcal{T}$ since it has the maximum value in the array of the root probabilities. Next in Step 1, the algorithm searches for the maximum probability value in the row $s_1$ while the column $s_1$ is excluded. The maximum value is this block is $0.4$ and corresponds to entry ($s_1$,$s_2$). Therefore, the sentence $s_2$ is added as the child node of the sentence $s_1$. In a similar fashion, the algorithm continues until all $6$ sentences are added to the tree.

\subsection{Structural Document-level Representation}
\label{sec:doc-rep}
We use a similar method presented in~\cite{Yang-Lapata2017} to extract a structurally rich representation for the entire document. First, for each sentence  (i.e., a discourse unit), we obtain a structurally-aware representation. To achieve this, we take into account the parent-child probabilities as well as the root probabilities as follows.
\begin{equation}
\begin{aligned}
\mathbf{p}_j &=  r_j \times\mathbf{e}_{root}+ \sum_{z=1}^{k}{\mathbf{A}[z,j] \times f_{z}}  \\
\mathbf{c}_{j} &= \sum_{z=1}^{k} {\mathbf{A}[j,z] \times f_{j}}  \\
\mathbf{g}_{j} &= \mathcal{G}(\mathbf{W} [\mathbf{p}_{j} || \mathbf{c}_{j} || f_{j} ]+\mathbf{b})
\end{aligned}
\end{equation}
\noindent where $\mathbf{p}_{j}$ and $\mathbf{c}_{j}$ are two context vectors  taking into account possible parents and children of a sentence $s_j$, respectively, $\mathbf{e}_{root}$ denotes a special root embedding vector, $||$ denotes vector concatenation operator, and $\mathbf{g}_{j}$ is a \emph{structurally-aware representation for sentence  $s_j$}. Finally, to extract a structurally rich representation for the entire document, we average all of its $\mathbf{g}_{j}$ vectors:
\begin{equation}
\begin{aligned}
x =\frac{1}{k} \sum_{j=1}^{k} \mathbf{g}_{j} \quad 
\label{eq:xi}
\end{aligned}
\end{equation}
\noindent where $x$ denotes \emph{the structurally rich document-level representation}.

\subsection{Fake News Classification}
\label{sec:fake-classification}
We hypothesize that the document-level structurally rich representations (Eq.~\ref{eq:xi}) offer a discriminatory power to detect fake news documents. Therefore, as shown in Figure~\ref{fig:model}, we employ a binary classification for fake news detection formulated as follows.
\begin{equation}
\begin{aligned}
c_i &= - \big( y_i  log(p^i_{f}) +(1-y_i) log(p^i_{r}) \big) \\
\mathcal{L(\theta)} &= \sum_{\forall d_i \in \mathcal{D}} c_i
\end{aligned}
\label{eq:loss}
\end{equation}
\noindent where $y_i$ is the ground-truth label of a document $d_i \in \mathcal{D}$, $c_i$ is the cross-entropy loss value, $p^i_{f}$ and $p^i_{r}$ are probabilities of $d_i$ being fake or real, respectively. $p^i_{f}$ and $p^i_{r}$ are obtained from the representation $x$  (Eq.~\ref{eq:xi}) fed to a fully connected layer followed by the softamx function. In Eq.~\ref{eq:loss}, $\theta$ denotes the framework's parameters and $\mathcal{L}$ is the total loss of the model. Since the entire framework (Eq.~\ref{eq:blstm} to \ref{eq:loss}) is fully differentiable, we utilize the error backpropagation~\cite{hecht1992theory} to calculate the gradients followed by the stochastic gradient descent~\cite{bottou2010large} to update and optimize the parameters. Note that the discourse dependency tree construction algorithm presented in Section~\ref{sec:tree-const} is employed post hoc. Hence, the gradient calculation over $argmax$ operator (line~\ref{alg:pc} of Algorithm~\ref{alg:tree-const}) is not performed.

%% file: structure-property.tex
\section{Structure-related Properties }
\label{sec:properties}
Incorporating discourse-level structure offer a discriminatory power distinguishing fake and real documents, which will be verified in Section~\ref{sec:comparison}.  
We expect more than this discriminatory power. We expect to identify insightful and interpretable information from extracted structures which can delineate intrinsic differences between fake and real news documents. To meet this expectation, we define three fundamental properties of constructed discourse dependency trees. Note that we leave the definitions of more advanced properties as one future work.  We seek to fulfill two goals by defining the structure-related properties. First, we intend to highlight the way that fake and real news documents are different.  Second, we intend to leverage these properties to shed light on the \textit{coherence} of fake and real news documents. Coherence has been the subject of many studies~\cite{Barzilay-etal2008,Lin-etal2011,Guinaudeau-Strube2013,Li-Hovy2014} and is concerned with how constituents of a document (e.g., discourse units) are linked together in a way that the entire text creates a clear mental picture to the readers~\cite{Storrer2002}. Notwithstanding its importance, coherence has not been investigated in the fake news domain in a large-scale and systematic fashion. Aiming at filling this gap, we naturally connect the defined structure-related properties with the coherence of news documents.

\begin{Property}[Number of Leaf Nodes]
This property, denoted as $\mathsf{P}_l$, defines the normalized number of leaf nodes in a discourse dependency tree: 
\label{property:leaves}
\end{Property} 
\begin{align}
\mathsf{P}_l = \frac{l}{{log}(k)} 
\end{align}
\noindent where $l$ is the number of leaf nodes (i.e., sentences) in the discourse dependency tree of a document. Recall that $k$ is the total number of sentences in a document. 

The intuition behind defining Property~\ref{property:leaves} is as follows. According to the description of the dependency tree in Section~\ref{sec:model-struct-extract}, leaf nodes are isolated discourse units and no other discourse units depend on them. Thus, the more the number of leaf nodes is, the less inter-linked the discourse units will be, and vice versa. Therefore, Property~\ref{property:leaves} is likely to indicate the coherence of a document -- the higher $\mathsf{P}_l$, the more isolated sentences and the less coherent the document. Also, for a document with $k$ sentences $\mathsf{P}_l \in [\frac{1}{log(k)},\frac{k-1}{log(k)}]$.

\begin{Property}[Preorder Difference]
This property, denoted as $\mathsf{P}_t$, defines the normalized positional  difference between the preorder traversal of a document's discourse dependency tree and its original sentential sequential order:   
\label{property:permuation}
\end{Property}
\begin{align}
\mathsf{P}_t = \frac{\sum_{j=1}^{k}{|s^{position}_j-j|}}{log(k)} 
\end{align} 
\noindent where $s^{position}_j$ denotes the position of a sentence $s_j$ in the preorder traversal\footnote{The subtrees are ordered based on \emph{when} they are added as the child nodes of a parent node in Algorithm~\ref{alg:tree-const}. That is why we can compute preorder traversal.} of dependency tree of a document e.g., $s^{position}_3=4$ in Figure~\ref{fig:dep-tree-example}. The position of $s_j$ in the original sequential order is simply $j$\footnote{We assume that the sentences in the sequential order are numbered incrementally from $1$ to $k$.}. The preorder traversal of the tree in Figure~\ref{fig:dep-tree-example} is the sequence $\{s_1, s_2, s_4, s_3,s_5\}$  and the sentential sequential order is $\{s_1,s_2,s_3,s_4,s_5\}$. Therefore, according to the definition of Property~\ref{property:permuation}:

\noindent $\mathsf{P}_t =\frac{|1-1|+|2-2|+|4-3|+|3-4|+|5-5|}{log(5)}\approx 2.86$.

The preorder traversal of a document's discourse dependency tree takes into consideration the organization of a document respect to the dependencies between sentences. Then, the purpose of Property~\ref{property:permuation} is to measure how much the organization of a document, captured through the preorder traversal, deviates from its sentential sequential order. Sentence order is highly related to the coherence of a document where the displaced order of sentences in a document makes it less coherent~\cite{Li-Hovy2014}. Thus, intuitively, the less the value of Property~\ref{property:permuation} for a document is, the more coherent that document should be. Also, for a document with $k$ sentences $\mathsf{P}_t \in [\frac{k-1}{log(k)},\frac{(k^2-1)}{2log(k)}]$ if $k$ is odd and $\mathsf{P}_t \in [\frac{k-1}{log(k)}, \frac{k^2}{2log(k)} ]$ if $k$ is even.

\begin{Property}[Parent-Child Distance]
This property, denoted as $\mathsf{P}_c$, defines the normalized sum of positional distances between child nodes and their parents when they are considered in the original sequential order:
\label{property:distance}
\end{Property}
\begin{align}
\mathsf{P}_c = \frac{\sum_{\forall c,p \in \mathcal{T}} |c_{position}-p_{position}|}{log(k)}
\end{align}
\noindent where $c_{position}$ and $p_{position}$ denote the positions of a child node $c$ and a parent node $p$, respectively, in the original sentential sequential order. For instance, in our running example, the parent node $s_3$ has $p_{position}=3$ (i.e., it is the third sentence) and its child node $s_5$ has $c_{position}=5$. Therefore, their parent-child distance is $|5-3|=2$. Following a similar calculation for other parent-child pairs, we have $\mathsf{P}_c=\frac{1+2+2+2}{log(5)} \approx 10$ .

Similar to Property~\ref{property:permuation}, Property~\ref{property:distance} pertains to the organization of a document and takes into consideration the deviation from sentential sequential order. Intuitively speaking, usually, we expect that a child node and its parent to be close to each other in the original sequential order. Consequently, the less value of this property is, the more coherent a document is likely to be. The range of Property~\ref{property:distance} in a document containing $k$ sentences is
$\mathsf{P}_c \in [\frac{k-1}{log(k)},\frac{k(k-1)}{2log(k)} ]$.

%% file: experiments.tex
\section{Experiments}
\label{sec:experiments}
To verify the performance of the proposed framework HDSF, we conduct a set of experiments. We seek to answer the following research questions:

\textbf{(1)}. How does the proposed framework perform on fake news detection?

\textbf{(2)}.  How do the defined structure-related properties describe the fake and real news documents?

In this section, we first describe the datasets followed by presenting the experimental settings. Afterward, we evaluate the performance of HDSF compared to several representative baselines. Finally, we present a structural analysis of the fake/real news documents.

\subsection{Datasets}
We utilize five available fake news datasets in this study. The first two datasets are collected by~\cite{Shu-etal2017b} and include online articles whose veracities have been identified by experts in BuzzFeed\footnote{\url{https://www.buzzfeed.com}} and PolitiFact\footnote{\url{http://www.politifact.com/}}. For the next two datasets, we utilize two available online fake news datasets provided by \textit{kaggle.com}\footnote{\url{https://www.kaggle.com/mrisdal/fake-news/data}} \footnote{\url{https://www.kaggle.com/jruvika/fake-news-detection}}. Finally, we include the dataset constructed and shared by McIntire\footnote{\url{https://github.com/GeorgeMcIntire/fake_real_news_dataset}}. Since the proposed framework HDSF is a general-purpose framework investigating discourse-level structures of fake/real news documents based on their textual contents, we do not restrict HDSF to a particular source of data and therefore combine all datasets. Similar to the previous work~\cite{Shu-etal2017b}, we balance the dataset to avoid a trivial solution as well as ensuring a fair performance comparison. In total, we have $3360$ fake and $3360$ real documents.

\subsection{Experimental Settings}
\label{sec:exp-settings}
First, we pre-process the documents by removing numbers, non-English characters, stop-words (e.g., `with'), and converting all characters to lower case. We randomly select $134$ documents as the development set, ($67$ from each class) and $134$ documents ($67$ from each class) as the test set. The remaining $6452$ documents are used for training. The development set is used for tuning the hyper-parameters.  We initialize the word embeddings from the Google news pre-trained word2vec embeddings~\cite{Mikolov-etal2013}\footnote{Out-of-vocabulary words are initialized randomly.}. LeakyReLU~\cite{Xu-etal2015} is used as the non-linear activation function and the number of hidden units in the BLSTM network is set to $100$. Each simulation is run for $200$ steps with a random mini-batch size of $40$ documents. The learning rate starts at $0.01$ with the decay rate of $0.9$ after every $50$ steps. We use the ADAM optimizer~\cite{Kingma-etal2014} to optimize the parameters. The PyTorch package\footnote{\url{https://pytorch.org/}} is utilized for the implementation and the code and data are publicly available in \url{https://github.com/hamidkarimi/HDSF}.

%% file: comparison.tex
\subsection{Comparison Results}
\label{sec:comparison}
To answer the research question (1), we compare the performance of HDSF with the following representative baselines.

\textbf{N-grams}. In this baseline method, we extract and combine unigrams, bigrams, and trigrams features and use SVM (Support Vector Machines)~\cite{scholkopf2001learning} for classification. 
 
\textbf{LIWC}~\cite{Pennebaker-etal2015}. LIWC (Linguistic Inquiry and
Word Count) offers a set of rich psycholinguistic features for a written document. We extract $94$ features for each document and use SVM for classification.
 
\textbf{RST}~\cite{Rubin-Lukoianova2015}. We extract a set of RST relations~\cite{Mann-Thompson1988} using the implementation of the method proposed by~\cite{ji2014representation}. Then, we vectorize the relations and employ SVM for classification. This baseline takes into account the hierarchical structure of documents via RST. 
 
\textbf{BiGRNN-CNN}~\cite{Ren-Zhang2016}. A CNN (Convolutional Neural Network) is applied at the sentence-level on word embeddings and a BiGRNN (Bi-Directional Gated Recurrent Neural Network) extracts features from a sequence of extracted sentential features. This baseline takes into consideration a two-level sequential structure for a document.
 
\textbf{LSTM[w+s]}. In this baseline, we apply an LSTM network on a sequence of word embeddings belonging to a sentence and then apply another LSTM on a sequence of extracted sentential features. LSTM[w+s] also considers a two-level sequential structure for a document. 

 \textbf{LSTM[s]}. This method is similar to LSTM[w+s] except that the mean of word embeddings in a sentence is used instead of applying an LSTM network. LSTM[s] considers a single sequential structure for a document.

We use accuracy as the metric of performance evaluation given that the dataset is fully balanced. Table~\ref{tab:comparison} shows the comparison results on the test set and we make the following observations:

\begin{itemize}
\item N-grams achieve a better performance than LIWC. In line with the previous study~\cite{Ott-etal2011}, this shows that for fake news detection, taking into account the context of a document as n-grams do is more effective than employing the existing pre-defined dictionaries as LIWC does.

\item Most of the time, methods wherein a document's structure is somehow taken into account outperform n-grams and LIWC. This observation shows that for fake news detection, the content's structure plays an important role.

\item The poor performance of RST is because of the following reasons: a) using RST without an annotated corpus is not very effective, and b) RST relations are extracted using auxiliary tools optimized for other corpora which cannot be applied effectively to the fake news corpus in hand. Note that annotating RST for our corpus is extremely unscalable and time-consuming. 

\item The proposed framework HDSF significantly outperforms all other methods. This observation shows that hierarchical discourse-level representations  are effectively rich for fake news prediction.

\end{itemize}

\begin{table}\small

\centering
\caption{Comparison results}
\begin{tabular}{c|c}
\hline
 Method& Accuracy (\%) \\
\hline 
N-grams &72.37\\ 
LIWC &70.26\\
RST & 67.68 \\
BiGRNN-CNN &77.06\\
LSTM[w+s]  &80.54\\
LSTM[s] &73.63\\
HDSF& \textbf{82.19} \\\hline 
\end{tabular}
\label{tab:comparison}
\end{table} 

\subsection{The Inspection of HDSF}

\begin{figure}[t!]
    \centering
    \begin{subfigure}[b]{0.25\textwidth}
        \includegraphics[scale=0.18]{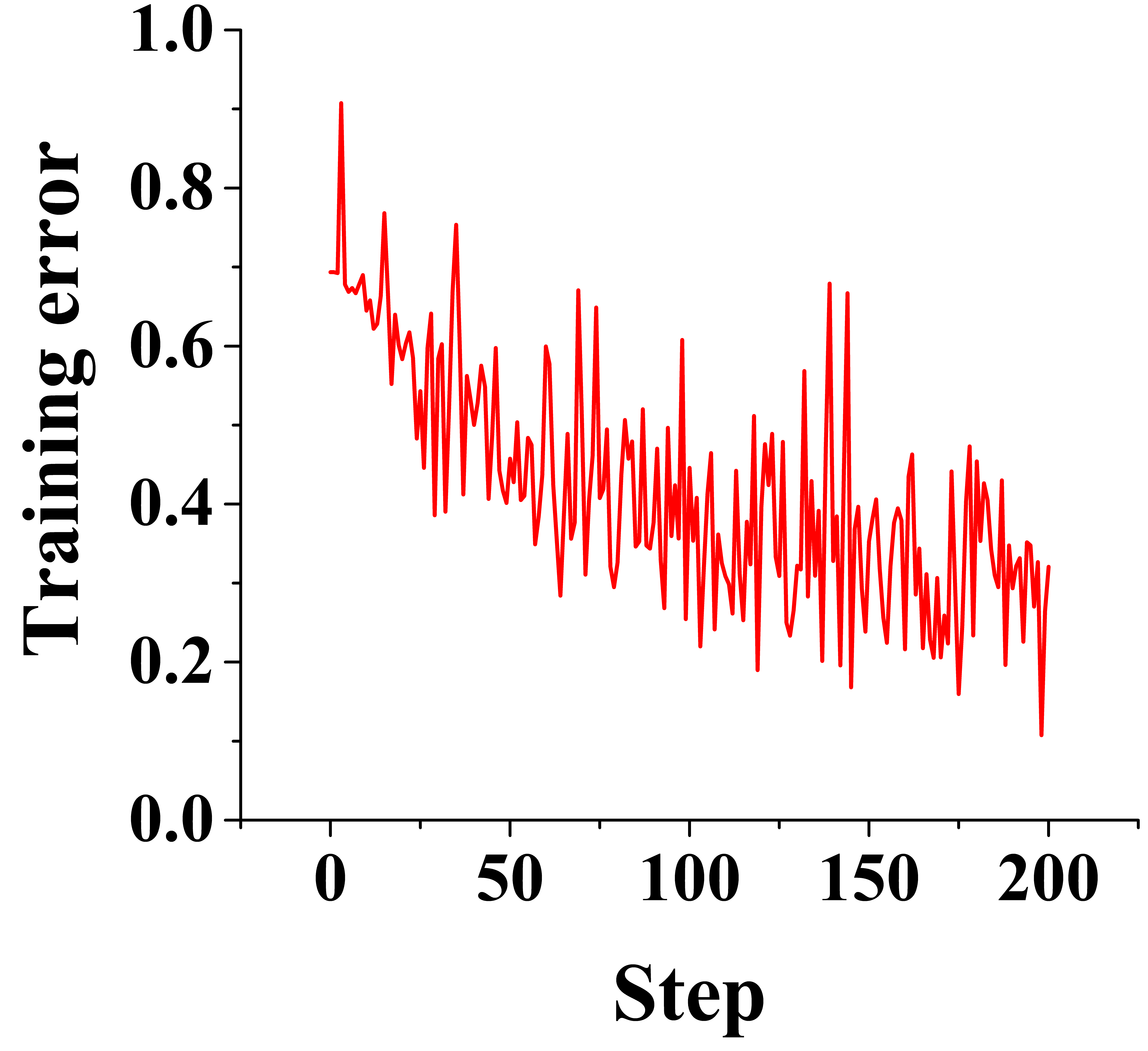}
        \caption{The training error}
        \label{subfig:curve}
    \end{subfigure}%
     ~
    \begin{subfigure}[b]{0.25\textwidth}
        \includegraphics[scale=0.18]{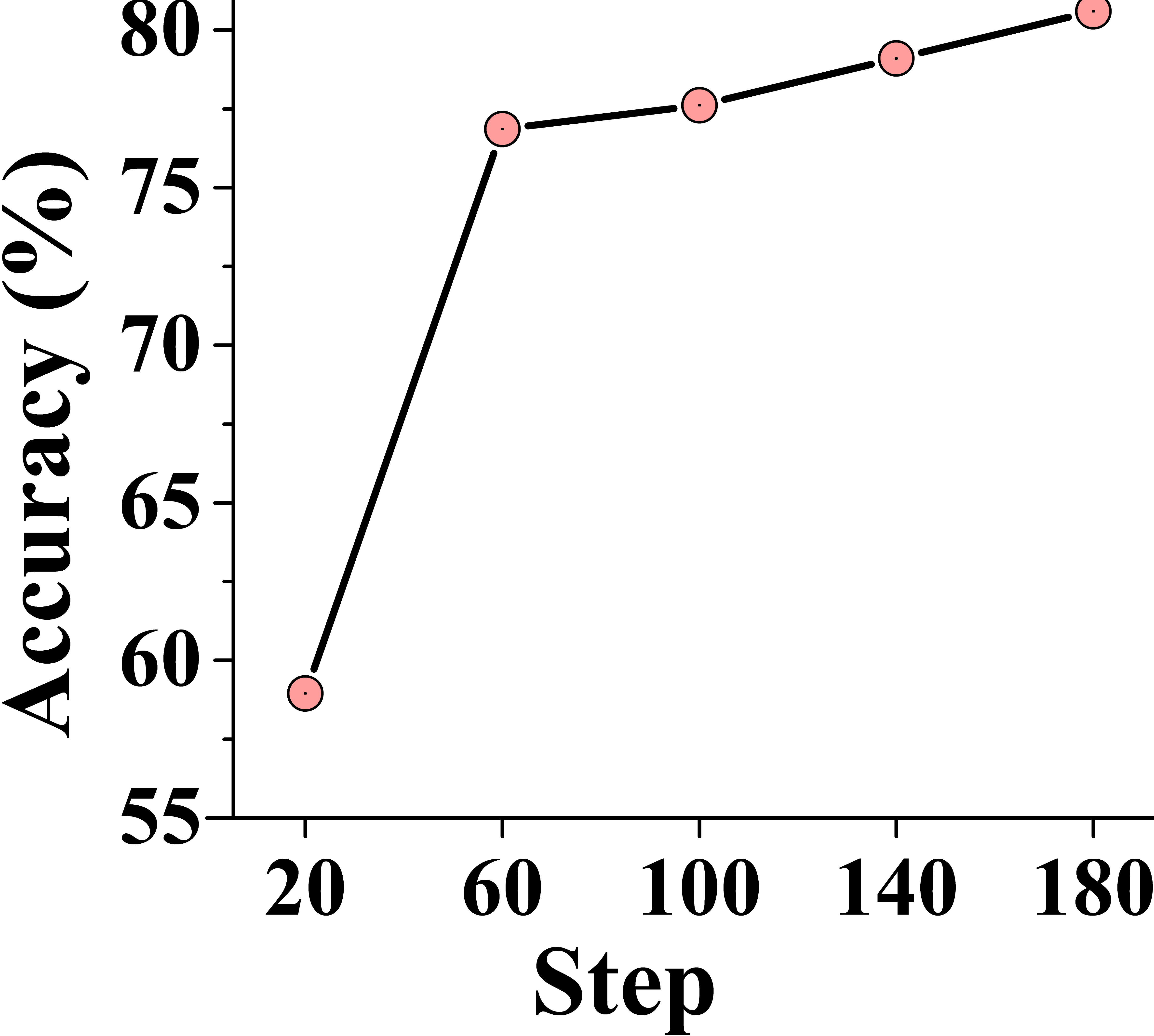}
        \caption{The accuracy on dev set}
        \label{subfig:dev}
    \end{subfigure}
    \caption{The Inspection of HDSF}
    \label{fig:inspection}
\end{figure}

To further verify the working of the HDSF framework, we inspect HDSF in more detail. Figure~\ref{subfig:curve} demonstrates the training error during model optimization. As we can observe from this figure, the error is decreasing as the training process proceeds. Furthermore, Figure~\ref{subfig:dev} demonstrates the accuracy on the development set during the training and it is monotonically increasing as the training goes on. Hence, based on these figures, we can ensure the framework is getting optimized and \emph{learns} to classify fake news documents correctly.

%% file: structure-exp.tex
\subsection{Structural Analysis for Fake/Real Documents}
\label{sec:structure}

\begin{figure}
\includegraphics[width=\columnwidth]{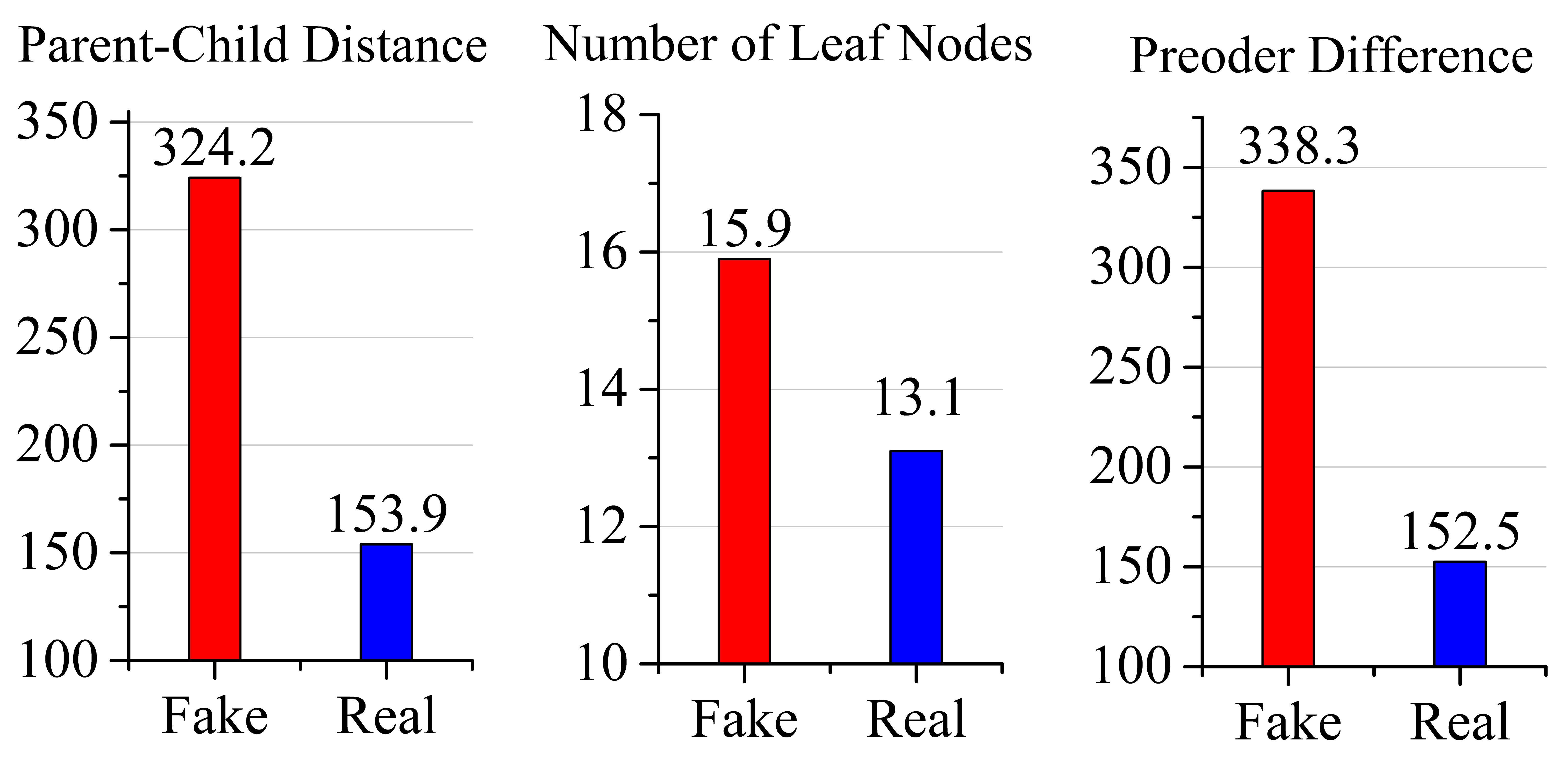}
\caption{Average values of the proposed structure-related properties for fake and real news documents}\label{fig:properties}

\end{figure}

In this section, we compute the average values of structure-related  properties, presented in Section~\ref{sec:properties}, for the fake/real news documents belonging to the test set. Figure~\ref{fig:properties} shows the results. We make the two key observations based on this figure:
\begin{itemize}
\item There is a significant difference in all three properties for fake news documents vs. real news documents. \emph{This observation shows the fact that structures of fake news documents at the discourse-level are substantially different from those of real ones}.
\item Noticeably, in all three properties, the real news documents show less value than the fake news documents. As described in Section~\ref{sec:properties}, all three properties are closely connected to the coherence of a document. \emph{ Therefore, real news documents indicate more degree of coherence}. 
\end{itemize} 

%% file: related.tex
\section{Related Work}
\label{sec:related}
Content-based fake news detection has been the subject of many linguistic research endeavors. \cite{DePaulo-etal2003} investigated fake stories introduced insightful cues in fake stories and highlighted `unusual' language in such stories. N-grams and Part-of-Speech (POS) tags are fundamental features of a text which have been utilized for fake news detection~\cite{Hadeer-etal2017,Ott-etal2013b}. Also, LIWC~\cite{Pennebaker-etal2015} has been employed to investigate the role of individual words in a document for deception detection~\cite{Ott-etal2011}. 
Since POS tags, n-grams, and LIWC features are considered as `shallow' and hand-crafted features,  deep neural networks have been utilized for fake news detection where features are extracted automatically~\cite{Wang2017,Ren-Zhang2016,Svitlana-etal2017,karimi2018multi}. In this study, we also utilized an automated feature extraction instead of relying on hand-craft features.  

Syntax-based approaches have been employed to take into account the hierarchical structure of textual documents for fake news detection~\cite{Feng-etal2012,Perez-Rosas-Mihalcea2015}. One caveat of syntax-based approaches is their reliance on auxiliary parsing tools which might propagate error in later part of a developed model. Moreover, generating syntactic production rules in an automated manner is a complicated process. 

Another way of incorporating structure is discourse-level parsing~\cite{Mann-Thompson1988,Li-etal2014,Ren-Zhang2016} which has seldom been explored for fake news detection. The noticeable exception is the approach developed by~\cite{Rubin-Lukoianova2015}. They extracted a set of RST relations in fake and real documents and vectorized them using a Vector Space Model (VSM) method~\cite{strehl2000impact}. Unlike \cite{Rubin-Lukoianova2015}, in this work, we proposed an automated and data-driven discourse-level parsing approach which used neither any annotated corpus nor any external tool.

%% file: NAACL2019Discourse.bbl
\begin{thebibliography}{38}
\expandafter\ifx\csname natexlab\endcsname\relax\def\natexlab#1{#1}\fi

\bibitem[{Ahmed et~al.(2017)Ahmed, Traore, and Saad}]{Hadeer-etal2017}
Hadeer Ahmed, Issa Traore, and Sherif Saad. 2017.
\newblock Detection of online fake news using n-gram analysis and machine
  learning techniques.
\newblock In \emph{International Conference on Intelligent, Secure, and
  Dependable Systems in Distributed and Cloud Environments}, pages 127--138.
  Springer.

\bibitem[{Allcott and Gentzkow(2017)}]{Allcott-Gentzkow2017}
Hunt Allcott and Matthew Gentzkow. 2017.
\newblock Social media and fake news in the 2016 election.
\newblock \emph{Journal of Economic Perspectives}, 31(2):211--36.

\bibitem[{Bachenko et~al.(2008)Bachenko, Fitzpatrick, and
  Schonwetter}]{Bachenko-etal2008}
Joan Bachenko, Eileen Fitzpatrick, and Michael Schonwetter. 2008.
\newblock Verification and implementation of language-based deception
  indicators in civil and criminal narratives.
\newblock In \emph{Proceedings of the 22nd International Conference on
  Computational Linguistics-Volume 1}, pages 41--48. Association for
  Computational Linguistics.

\bibitem[{Barzilay and Lapata(2008)}]{Barzilay-etal2008}
Regina Barzilay and Mirella Lapata. 2008.
\newblock Modeling local coherence: An entity-based approach.
\newblock \emph{Computational Linguistics}, 34(1):1--34.

\bibitem[{Bhatia et~al.(2015)Bhatia, Ji, and Eisenstein}]{bhatia2015better}
Parminder Bhatia, Yangfeng Ji, and Jacob Eisenstein. 2015.
\newblock Better document-level sentiment analysis from rst discourse parsing.
\newblock In \emph{Proceedings of the 2015 Conference on Empirical Methods in
  Natural Language Processing}, pages 2212--2218.

\bibitem[{Bottou(2010)}]{bottou2010large}
L{\'e}on Bottou. 2010.
\newblock Large-scale machine learning with stochastic gradient descent.
\newblock In \emph{Proceedings of COMPSTAT'2010}, pages 177--186. Springer.

\bibitem[{DePaulo et~al.(2003)DePaulo, Lindsay, Malone, Muhlenbruck, Charlton,
  and Cooper}]{DePaulo-etal2003}
Bella~M DePaulo, James~J Lindsay, Brian~E Malone, Laura Muhlenbruck, Kelly
  Charlton, and Harris Cooper. 2003.
\newblock Cues to deception.
\newblock \emph{Psychological bulletin}, 129(1):74.

\bibitem[{Feng et~al.(2012)Feng, Banerjee, and Choi}]{Feng-etal2012}
Song Feng, Ritwik Banerjee, and Yejin Choi. 2012.
\newblock Syntactic stylometry for deception detection.
\newblock In \emph{Proceedings of the 50th Annual Meeting of the Association
  for Computational Linguistics: Short Papers-Volume 2}, pages 171--175.
  Association for Computational Linguistics.

\bibitem[{Guinaudeau and Strube(2013)}]{Guinaudeau-Strube2013}
Camille Guinaudeau and Michael Strube. 2013.
\newblock Graph-based local coherence modeling.
\newblock In \emph{Proceedings of the 51st Annual Meeting of the Association
  for Computational Linguistics (Volume 1: Long Papers)}, volume~1, pages
  93--103.

\bibitem[{Hecht-Nielsen(1992)}]{hecht1992theory}
Robert Hecht-Nielsen. 1992.
\newblock Theory of the backpropagation neural network.
\newblock In \emph{Neural networks for perception}, pages 65--93. Elsevier.

\bibitem[{Hochreiter and Schmidhuber(1997)}]{Hochreiter-Schmidhuberl1997}
Sepp Hochreiter and J{\"u}rgen Schmidhuber. 1997.
\newblock Long short-term memory.
\newblock \emph{Neural computation}, 9(8):1735--1780.

\bibitem[{Ji and Eisenstein(2014)}]{ji2014representation}
Yangfeng Ji and Jacob Eisenstein. 2014.
\newblock Representation learning for text-level discourse parsing.
\newblock In \emph{Proceedings of the 52nd Annual Meeting of the Association
  for Computational Linguistics (Volume 1: Long Papers)}, volume~1, pages
  13--24.

\bibitem[{Karimi et~al.(2018)Karimi, Roy, Saba-Sadiya, and
  Tang}]{karimi2018multi}
Hamid Karimi, Proteek Roy, Sari Saba-Sadiya, and Jiliang Tang. 2018.
\newblock Multi-source multi-class fake news detection.
\newblock In \emph{Proceedings of the 27th International Conference on
  Computational Linguistics}, pages 1546--1557.

\bibitem[{Kim et~al.(2017)Kim, Denton, Hoang, and Rush}]{Kim-etal2017}
Yoon Kim, Carl Denton, Luong Hoang, and Alexander~M Rush. 2017.
\newblock Structured attention networks.
\newblock \emph{arXiv preprint arXiv:1702.00887}.

\bibitem[{Kingma and Ba(2014)}]{Kingma-etal2014}
Diederik~P Kingma and Jimmy Ba. 2014.
\newblock Adam: A method for stochastic optimization.
\newblock \emph{arXiv preprint arXiv:1412.6980}.

\bibitem[{Li and Hovy(2014)}]{Li-Hovy2014}
Jiwei Li and Eduard Hovy. 2014.
\newblock A model of coherence based on distributed sentence representation.
\newblock In \emph{Proceedings of the 2014 Conference on Empirical Methods in
  Natural Language Processing (EMNLP)}, pages 2039--2048.

\bibitem[{Li et~al.(2014{\natexlab{a}})Li, Li, and Hovy}]{Li-etal2014}
Jiwei Li, Rumeng Li, and Eduard Hovy. 2014{\natexlab{a}}.
\newblock Recursive deep models for discourse parsing.
\newblock In \emph{Proceedings of the 2014 Conference on Empirical Methods in
  Natural Language Processing (EMNLP)}, pages 2061--2069.

\bibitem[{Li et~al.(2014{\natexlab{b}})Li, Wang, Cao, and Li}]{Li-etal2014b}
Sujian Li, Liang Wang, Ziqiang Cao, and Wenjie Li. 2014{\natexlab{b}}.
\newblock Text-level discourse dependency parsing.
\newblock In \emph{Proceedings of the 52nd Annual Meeting of the Association
  for Computational Linguistics (Volume 1: Long Papers)}, volume~1, pages
  25--35.

\bibitem[{Lin et~al.(2011)Lin, Ng, and Kan}]{Lin-etal2011}
Ziheng Lin, Hwee~Tou Ng, and Min-Yen Kan. 2011.
\newblock Automatically evaluating text coherence using discourse relations.
\newblock In \emph{Proceedings of the 49th Annual Meeting of the Association
  for Computational Linguistics: Human Language Technologies-Volume 1}, pages
  997--1006. Association for Computational Linguistics.

\bibitem[{Liu and Lapata(2018)}]{Yang-Lapata2017}
Yang Liu and Mirella Lapata. 2018.
\newblock Learning structured text representations.
\newblock \emph{Transactions of the Association of Computational Linguistics},
  6:63--75.

\bibitem[{Mann and Thompson(1988)}]{Mann-Thompson1988}
William~C Mann and Sandra~A Thompson. 1988.
\newblock Rhetorical structure theory: Toward a functional theory of text
  organization.
\newblock \emph{Text-Interdisciplinary Journal for the Study of Discourse},
  8(3):243--281.

\bibitem[{Mikolov et~al.(2013)Mikolov, Chen, Corrado, and
  Dean}]{Mikolov-etal2013}
Tomas Mikolov, Kai Chen, Greg Corrado, and Jeffrey Dean. 2013.
\newblock Efficient estimation of word representations in vector space.
\newblock \emph{arXiv preprint arXiv:1301.3781}.

\bibitem[{Morey et~al.(2018)Morey, Muller, and Asher}]{Mathieu-etal2018}
Mathieu Morey, Philippe Muller, and Nicholas Asher. 2018.
\newblock A dependency perspective on rst discourse parsing and evaluation.
\newblock \emph{Computational Linguistics}, (Just Accepted):1--54.

\bibitem[{Ott et~al.(2013)Ott, Cardie, and Hancock}]{Ott-etal2013b}
Myle Ott, Claire Cardie, and Jeffrey~T Hancock. 2013.
\newblock Negative deceptive opinion spam.
\newblock In \emph{Proceedings of the 2013 conference of the north american
  chapter of the association for computational linguistics: human language
  technologies}, pages 497--501.

\bibitem[{Ott et~al.(2011)Ott, Choi, Cardie, and Hancock}]{Ott-etal2011}
Myle Ott, Yejin Choi, Claire Cardie, and Jeffrey~T Hancock. 2011.
\newblock Finding deceptive opinion spam by any stretch of the imagination.
\newblock In \emph{Proceedings of the 49th Annual Meeting of the Association
  for Computational Linguistics: Human Language Technologies-Volume 1}, pages
  309--319. Association for Computational Linguistics.

\bibitem[{Pennebaker et~al.(2015)Pennebaker, Boyd, Jordan, and
  Blackburn}]{Pennebaker-etal2015}
James~W Pennebaker, Ryan~L Boyd, Kayla Jordan, and Kate Blackburn. 2015.
\newblock The development and psychometric properties of liwc2015.
\newblock Technical report.

\bibitem[{P{\'e}rez-Rosas and Mihalcea(2015)}]{Perez-Rosas-Mihalcea2015}
Ver{\'o}nica P{\'e}rez-Rosas and Rada Mihalcea. 2015.
\newblock Experiments in open domain deception detection.
\newblock In \emph{Proceedings of the 2015 Conference on Empirical Methods in
  Natural Language Processing}, pages 1120--1125.

\bibitem[{Prasad et~al.(2007)Prasad, Miltsakaki, Dinesh, Lee, Joshi, Robaldo,
  and Webber}]{Prasad-etal2007}
Rashmi Prasad, Eleni Miltsakaki, Nikhil Dinesh, Alan Lee, Aravind Joshi, Livio
  Robaldo, and Bonnie~L Webber. 2007.
\newblock The penn discourse treebank 2.0 annotation manual.

\bibitem[{Ren and Zhang(2016)}]{Ren-Zhang2016}
Yafeng Ren and Yue Zhang. 2016.
\newblock Deceptive opinion spam detection using neural network.
\newblock In \emph{Proceedings of COLING 2016, the 26th International
  Conference on Computational Linguistics: Technical Papers}, pages 140--150.

\bibitem[{Rubin and Lukoianova(2015)}]{Rubin-Lukoianova2015}
Victoria~L Rubin and Tatiana Lukoianova. 2015.
\newblock Truth and deception at the rhetorical structure level.
\newblock \emph{Journal of the Association for Information Science and
  Technology}, 66(5):905--917.

\bibitem[{Scholkopf and Smola(2001)}]{scholkopf2001learning}
Bernhard Scholkopf and Alexander~J Smola. 2001.
\newblock \emph{Learning with kernels: support vector machines, regularization,
  optimization, and beyond}.
\newblock MIT press.

\bibitem[{Schuster and Paliwal(1997)}]{Schuster-Paliwal1997}
Mike Schuster and Kuldip~K Paliwal. 1997.
\newblock Bidirectional recurrent neural networks.
\newblock \emph{IEEE Transactions on Signal Processing}, 45(11):2673--2681.

\bibitem[{Shu et~al.(2017)Shu, Wang, and Liu}]{Shu-etal2017b}
Kai Shu, Suhang Wang, and Huan Liu. 2017.
\newblock Exploiting tri-relationship for fake news detection.
\newblock \emph{arXiv preprint arXiv:1712.07709}.

\bibitem[{Storrer(2002)}]{Storrer2002}
Angelika Storrer. 2002.
\newblock Coherence in text and hypertext.
\newblock \emph{Document Design}, 3(2):156--168.

\bibitem[{Strehl and Ghosh(2000)}]{strehl2000impact}
Alexander Strehl and Joydeep Ghosh. 2000.
\newblock Impact of similarity measures on web-page clustering.

\bibitem[{Volkova et~al.(2017)Volkova, Shaffer, Jang, and
  Hodas}]{Svitlana-etal2017}
Svitlana Volkova, Kyle Shaffer, Jin~Yea Jang, and Nathan Hodas. 2017.
\newblock Separating facts from fiction: Linguistic models to classify
  suspicious and trusted news posts on twitter.
\newblock In \emph{Proceedings of the 55th Annual Meeting of the Association
  for Computational Linguistics (Volume 2: Short Papers)}, volume~2, pages
  647--653.

\bibitem[{Wang(2017)}]{Wang2017}
William~Yang Wang. 2017.
\newblock " liar, liar pants on fire": A new benchmark dataset for fake news
  detection.
\newblock \emph{arXiv preprint arXiv:1705.00648}.

\bibitem[{Xu et~al.(2015)Xu, Wang, Chen, and Li}]{Xu-etal2015}
Bing Xu, Naiyan Wang, Tianqi Chen, and Mu~Li. 2015.
\newblock Empirical evaluation of rectified activations in convolutional
  network.
\newblock \emph{arXiv preprint arXiv:1505.00853}.

\end{thebibliography}
